\documentclass[10pt,twocolumn,letterpaper]{article}

\usepackage[pagenumbers]{cvpr}         
\usepackage[utf8]{inputenc}
\usepackage{amsmath,amssymb,amsfonts}
\usepackage{graphicx}
\usepackage{textcomp}
\usepackage{subcaption}
\usepackage{xcolor}
\usepackage{url}
\usepackage{algorithm}
\usepackage{algpseudocode}
\usepackage{multirow}
\usepackage{booktabs}

\definecolor{cvprblue}{rgb}{0.21,0.49,0.74}
\usepackage[pagebackref,breaklinks,colorlinks,allcolors=cvprblue]{hyperref}

\title{DD-CAM: Minimal Sufficient Explanations for Vision Models Using Delta Debugging}

\author{
\begin{minipage}{0.45\linewidth}
\centering
\textbf{Krishna Khadka}\\
Department of Computer Science \& Engineering\\
The University of Texas at Arlington\\
Arlington, Texas, USA\\
{\tt\small krishna.khadka@mavs.uta.edu}
\end{minipage}
\hfill
\begin{minipage}{0.45\linewidth}
\centering
\textbf{Yu Lei}\\
Department of Computer Science \& Engineering\\
The University of Texas at Arlington\\
Arlington, Texas, USA\\
{\tt\small ylei@cse.uta.edu}
\end{minipage}
\\[3.5em]  
\begin{minipage}{0.45\linewidth}
\centering
\textbf{Raghu N. Kacker}\\
Information Technology Laboratory\\
National Institute of Standards and Technology (NIST)\\
Gaithersburg, Maryland, USA\\
{\tt\small raghu.kacker@nist.gov}
\end{minipage}
\hfill
\begin{minipage}{0.45\linewidth}
\centering
\textbf{D. Richard Kuhn}\\
Information Technology Laboratory\\
National Institute of Standards and Technology (NIST)\\
Gaithersburg, Maryland, USA\\
{\tt\small d.kuhn@nist.gov}
\end{minipage}
}

\begin{document}
\maketitle

\begin{abstract}
We introduce a gradient-free framework for identifying minimal, sufficient, and decision-preserving explanations in vision models by isolating the smallest subset of representational units whose joint activation preserves predictions. Unlike existing approaches that aggregate all units, often leading to cluttered saliency maps, our approach, DD-CAM, identifies a 1-minimal subset whose joint activation suffices to preserve the prediction (i.e., removing any unit from the subset alters the prediction). To efficiently isolate minimal sufficient subsets, we adapt delta debugging, a systematic reduction strategy from software debugging, and configure its search strategy based on unit interactions in the classifier head: testing individual units for models with non-interacting units and testing unit combinations for models in which unit interactions exist. We then generate minimal, prediction-preserving saliency maps that highlight only the most essential features. Our experimental evaluation demonstrates that our approach can produce more faithful explanations and achieve higher localization accuracy than the state-of-the-art CAM-based approaches.
\end{abstract} \vspace{-1.5em} 

\section{Introduction}
Deep Convolutional Neural Networks (CNNs) and Vision Transformers (ViTs) have achieved state-of-the-art performance in vision tasks such as image classification, object detection, and segmentation~\cite{krizhevsky2012imagenet, dosovitskiy2020image}. Despite this success, these models remain largely uninterpretable, raising concerns in high-stakes domains like healthcare, finance, and autonomous systems~\cite{arrieta2020explainable, holzinger2017we}.

Class Activation Mapping (CAM) is a widely used post-hoc technique that generates saliency maps by leveraging internal representations from vision models. CAM-based approaches use feature maps (in CNNs) or patch tokens (in ViTs), typically from the final layer before classification—because these representations balance high-level semantic content with spatial structure suitable for localization. Gradient-based CAM approaches like Grad-CAM~\cite{selvaraju2017grad} weight features using gradients, while gradient-free alternatives like Score-CAM~\cite{wang2020score} and Ablation-CAM~\cite{desai2020ablation} use forward-pass perturbations. However, all of the existing approaches aggregate contributions from every unit, often producing cluttered saliency maps that obscure which features are truly necessary for the prediction.

We address the above limitation by reframing explanation generation as identifying \textit{minimal sufficient subsets}: a smallest set of representational units whose joint activation preserves the model's prediction. A representational unit (feature map or patch) is necessary with respect to the set if removing it from the set changes the prediction; here, necessity is defined in the local sense, meaning the unit is required for this particular set to preserve the prediction. A set is sufficient if activating only that set preserves the prediction. This minimal sufficiency-based formulation ensures that each selected unit is locally necessary and produces focused explanations that highlight only the essential regions driving the prediction. Furthermore, it generalizes across architectures: whether a model uses convolutions or self-attention, we can uniformly ask which minimal subset of representational units preserves the prediction.

We adapt delta debugging, a systematic reduction strategy from software debugging originally designed to isolate minimal failure-inducing inputs~\cite{zeller2002simplifying}, to identify minimal sufficient subsets in vision models. In software debugging, the algorithm begins with a known failure and attempts to find a set of inputs that still preserves that failure. In our setting, the model's prediction plays the same role as the outcome to preserve; instead of preserving a program failure, we aim to preserve the model's prediction. Representational units (feature maps or patch tokens) serve as testable components. The algorithm recursively partitions candidate units and tests whether subsets preserve the prediction, ultimately identifying a 1-minimal set where removing any single unit changes the prediction.

We optimize the performance of the delta debugging algorithm based on the classifier head employed in a vision model. For models where units interact in the classifier head — such as CNNs with multiple fully connected (FC) layers and ReLU activations (e.g., VGG models~\cite{simonyan2014very}) or ViTs where self-attention creates dependencies between patches, the general delta debugging algorithm is used. For models where units do not interact in the classifier head, i.e., they contribute to the model prediction independently, such as CNNs with Global Average  Pooling (GAP) followed by a single fully connected layer (e.g., ResNet-50~\cite{he2016deep}, EfficientNet-B0~\cite{tan2019efficientnet}, Inception-v3~\cite{szegedy2016rethinking}), we configure the delta debugging algorithm to test each unit individually, i.e., without considering their combinations, reducing the computation complexity. Both configurations provide the same 1-minimality guarantee. 

We evaluated our approach by examining both explanation faithfulness and localization accuracy. For faithfulness, we used eight ImageNet-pretrained models~\cite{russakovsky2015imagenet}, including six CNNs and two vision transformers, and generated explanations for 2,000 ImageNet validation images~\cite{russakovsky2015imagenet}. We compared DD-CAM against seven state-of-the-art CAM-based methods. Our results show that DD-CAM outperforms all baselines in 15 out of 18 evaluations, averaged across model groups, providing more faithful explanations across both convolutional and transformer models. 

For localization accuracy, we conduct a separate evaluation on 1,000 radiologist-annotated chest X-rays from NIH ChestX-ray14~\cite{wang2017chestx} and compared to the same baselines. DD-CAM achieves significantly higher localization accuracy in terms of improving IoU by 45\% and precision by 22\% over the strongest baseline while producing compact single-region explanations.

Our contributions are threefold. (1) We introduce a gradient-free approach for identifying minimal decision-preserving representational units, providing the first application of delta debugging to vision model explanations. (2) We perform extensive evaluations to show that minimal sufficient explanations improve both faithfulness and localization. (3) We release an implementation, \emph{DD-CAM}, which is anonymously available for review~\cite{ddcam2025anonymous}\

\section{Related Work}
\label{sec:related}
We review post-hoc, attribution-based explanation methods, beginning with model-agnostic approaches, followed by two model-specific families relevant to our work: sensitivity-based explanations and CAM techniques.

\subsection{Model-Agnostic Explanations}
Model-agnostic methods such as LIME~\cite{ribeiro2016should}, SHAP~\cite{lundberg2017unified}, and related variants~\cite{ribeiro2018anchors,lundberg2018consistent} explain predictions by fitting surrogate models to perturbed inputs or by estimating Shapley values. Because they operate purely in the input space, they treat the network as a black box and rely on heuristic baselines (e.g., blur, zeros, noise) and masking schemes (e.g., superpixels), which can substantially affect explanation quality and stability~\cite{adebayo2018sanity}. In contrast, we work with internal activations and use zero-masking of representational units as a natural, architecture-consistent baseline.

\subsection{Sensitivity-Based Explanations}
Sensitivity-based methods—including Saliency Maps~\cite{simonyan2013deep}, Guided Backpropagation~\cite{springenberg2015striving}, Integrated Gradients~\cite{sundararajan2017axiomatic}, and SmoothGrad~\cite{smilkov2017smoothgrad}—attribute importance using input-output gradients. Although efficient, they often produce noisy attributions due to gradient locality and saturation in deep networks~\cite{adebayo2018sanity}, and they do not indicate which internal representations drive the prediction. In contrast, our gradient-free approach identifies minimal subsets of internal units that preserve the output, offering a more directly causal explanation.

\subsection{Class Activation Mapping Explanations}
CAM-based methods leverage internal activation patterns in the final layers of
vision models. Grad-CAM~\cite{selvaraju2017grad} and
Grad-CAM++~\cite{chattopadhay2018grad} use class-specific gradients to weight
representational units, but are affected by gradient saturation and noise.
Gradient-free variants such as Score-CAM~\cite{wang2020score} and
Ablation-CAM~\cite{desai2020ablation} estimate unit contributions through
multiple forward passes, improving robustness but still aggregating all units
into dense saliency maps. In contrast, we identify a minimal
decision-preserving subset that is sufficient to retain the prediction, yielding
stricter explanations focused on essential features. For Vision Transformers, attention-based explainers~\cite{chefer2021transformer,abnar2020quantifying}
derive relevance from attention weights, whereas our method again searches for
minimal sufficient subsets of patch tokens.

Minimal-input-region methods such as Sufficient Input Subsets
(SIS)~\cite{carter2019made} and Meaningful
Perturbation~\cite{fong2017interpretable} also pursue minimality but operate in the input space and do not produce attribution scores. SIS iteratively removes input features until the confidence drops, while Meaningful Perturbation learns a sparse, smooth mask. In contrast, we enforce minimality over internal representations and still generate attribution maps from the selected units.

\section{Background}
This section introduces internal representations in vision models and delta debugging from software engineering.

\subsection{Vision Model Representations}
\label{subsec:vision_representations}
Vision models for image classification share a common architecture: a feature extractor producing internal representations, followed by a classifier head mapping these to class predictions.

CNNs process images through convolutional layers, learning hierarchical visual features~\cite{krizhevsky2012imagenet}. The final convolutional layer produces $K$ feature maps $A_k \in \mathbb{R}^{H \times W}$ that retain spatial structure while encoding high-level semantic patterns.

Vision Transformers divide images into patches, embed them via linear projection, and process them through transformer blocks with self-attention~\cite{dosovitskiy2020image}. 
Following ViT-ReciproCAM~\cite{byun2023vit}, we extract patch tokens from the first LayerNorm layer of the final transformer encoder block, which produces $N$ patch tokens $P_n \in \mathbb{R}^D$ (typically 196 image patches plus one CLS token). This layer provides unmodified skip-connected features suitable for explainability.

Classifier heads map these representations to predictions. Linear heads use Global Average Pooling followed by a single fully connected layer, where each unit contributes independently. Non-linear heads use multiple layers with ReLU 
activations, creating interactions between units.

Across architectures, both CNNs and ViTs produce a set of representational units before classification: feature maps or patch tokens. We can uniformly ask: which subset of units is necessary for the prediction? 

\subsection{Delta Debugging}
\label{subsec:delta_debugging}
Delta debugging systematically isolates minimal failure-inducing input components in software debugging~\cite{zeller2002simplifying}. Given a program input causing failure, it identifies the smallest subset of components whose removal prevents the failure. The algorithm recursively partitions the input into $n$ subsets and tests each subset and its complement. If a subset or its complement reproduces the failure, the search continues within the subset or its complement. Otherwise, i.e., if none of the subsets or their complements reproduce the failure, the algorithm increases the partition granularity ($n \leftarrow 2n$) and repeats the same process. The algorithm terminates when a 1-minimal set is found, i.e., removing any single component prevents the failure.
\vspace{-0.8em} 


\section{Motivation}
\label{sec:motivation}
Our goal is to identify minimal sufficient explanations: the smallest subset of 
representational units (feature maps or patch tokens) whose joint activation preserves the model's prediction. A subset $S \subseteq \{1,\ldots,M\}$ is \textit{sufficient} if masking all units outside $S$ preserves the predicted class $\hat{c}$. It is \textit{1-minimal} if removing any single unit from $S$ changes the prediction. While 1-minimality does \emph{not} guarantee the smallest possible set (minimum cardinality), it provides a tractable approximation, as finding the absolute minimal set is NP hard~\cite{zeller2002simplifying}. If multiple 1-minimal subsets exist, we identify one of them and use it to generate explanations~\cite{zeller2002simplifying}.

We draw inspiration from fault localization in software debugging, where the goal is to isolate minimal failure-inducing inputs. In our setting, the prediction is 
the failure to preserve, representational units are testable components, and masking 
units (setting activations to zero) tests their removal. This analogy motivates 
adapting delta debugging to vision model explanation.

The concept of minimal sufficient explanation is agnostic to internal representations. Whether a model uses feature maps (CNNs) or patch tokens (ViTs), we can uniformly test subsets for prediction preservation. Only the tested units differ while the algorithmic logic remains unchanged. This enables a unified framework across architectures.

Minimal sufficient explanations offer several advantages: 
(1) reduced cognitive load by focusing on necessary features~\cite{miller1956magical, cowan2001magical}, (2) causal grounding since each unit is provably necessary, (3) targeted robustness checks in safety-critical applications~\cite{doshi2017towards}, and (4) visual clarity through less cluttered saliency maps.

\section{Approach}
\label{sec:approach}
Let $f : \mathbb{R}^{L \times H \times W} \to \mathbb{R}^{C}$ be a pretrained vision model, where $L$ is the number of input channels and $H \times W$ is the spatial resolution. An input image is denoted by $I \in \mathbb{R}^{L \times H \times W}$.
The model produces class scores $f(I) \in \mathbb{R}^{C}$, where $C$ is the number of classes and $y_c = f(I)_c$ is the logit for class $c$. The predicted label is $\hat{c} = \arg\max_c y_c$.

Our objective is to explain the model's decision $\hat{c}$ by identifying a 1-minimal subset of representational units that preserves this prediction. We target the final layer before classification because it balances high-level 
semantic abstraction with spatial detail, making it suitable for visual interpretability~\cite{zhou2016learning}. For CNNs, the final convolutional layer produces $A = [A_1, A_2, \dots, A_K] \in \mathbb{R}^{K \times h \times w}$, where 
$K$ is the number of feature maps. For ViTs, we test subsets of the $N$  
patch tokens at the input to the final transformer block, while the CLS token is 
always preserved. The patches are represented as $P = [P_1, P_2, \dots, P_N] \in \mathbb{R}^{N \times D}$, 
where $D$ is the embedding dimension. Our goal is to find $S^\star \subseteq \{1,\dots,M\}$ 
(where $M=K$ for CNNs or $M=N$ for ViTs) such that activating only the units indexed by $S^\star$ and zero-masking the rest preserves $\hat{c}$. Here, prediction preservation means that the top-1 predicted class remains unchanged. As discussed in Section \ref{subsec:subset_selection}, unlike input-space masking, zero-masking at the representation level is well-defined. 

We introduce a gradient-free approach based on delta debugging to identify $S^\star$. The approach is a three-stage pipeline: (1) Activation Extraction, (2) Subset Selection via Delta Debugging, and (3) Saliency Map Generation. 

\subsection{Step 1: Activation Extraction}
\label{subsec:activation_extraction}

Initially, a forward pass is performed with the input image $I$ to compute the 
original prediction $\hat{c}$ and the representational tensor from the target 
layer. For CNNs, we extract feature maps $A = [A_1, \dots, A_K] \in \mathbb{R}^{K \times h \times w}$ 
from the final convolutional layer using a forward hook. For Vision Transformers, 
we intervene on the $N$ patch tokens at the input to the final transformer 
block, following ViT-ReciproCAM~\cite{byun2023vit}. The CLS token is always 
preserved during masking operations to maintain skip-connection information and 
ensure proper attention mechanism function. We represent the testable patch tokens 
as $P = [P_1, \dots, P_N] \in \mathbb{R}^{N \times D}$.

We split the model at the target layer to reuse cached representations, enabling 
partial forward passes from the intervention point through the remaining layers. 
The extracted tensor and prediction $\hat{c}$ are cached for subsequent perturbation 
tests. We refer to this downstream subnetwork as the remainder network $f_{\text{rem}}$.

\subsection{Step 2: Subset Selection via Delta Debugging}
\label{subsec:subset_selection}
This stage identifies a minimal subset of representational units $S^\star$ using delta debugging. We employ perturbation-based testing~\cite{desai2020ablation}, selectively masking (zeroing out) units and observing if the model's prediction $\hat{c}$ remains unchanged. 

Unlike masking in the input space, where defining a neutral reference value is challenging, masking at the representation level uses zero as a natural reference value~\cite{desai2020ablation}, effectively removing the unit's activation. For CNNs, zeroing a feature map $A_k$ sets all 
activations to zero. For ViTs, masking a patch token $P_n$ sets all token dimensions to zero while preserving the CLS token, ensuring the final self-attention layer operates only on the selected patches. In both cases, zero represents the absence of activation, avoiding artificial patterns and providing 
a well-defined reference.

\subsubsection{Delta Debugging Algorithm}
\label{ssubsec:delta_debugging}
Algorithm~\ref{alg:dd} adapts the standard delta debugging (DD) procedure to identify a 1-minimal sufficient set of representational units. We begin with the full set $S = \{1,\dots,M\}$ and an initial granularity $n = 2$. 
In each recursive call, $S$ is partitioned into $n$ disjoint subsets $S_1,\dots,S_n$. For each subset $S_i$, the complement $S \setminus S_i$ is evaluated by zeroing the units in $S_i$ and performing a forward pass through $f_{\text{rem}}$. If the complement preserves the original prediction, then $S_i$ is unnecessary. In this case, delta debugging resets the granularity to $n = 2$ and recurses on the reduced set $S \setminus S_i$. If no subset can be removed at the current 
granularity, the granularity is increased to $\min(2n, |S|)$ and DD recurses on this larger value of $n$. When no further reduction is possible (that is, when $n = |S|$ and each single-element complement fails to preserve the 
prediction), the algorithm returns a 1-minimal sufficient set $S^\star$.

If units are interacting, masking one unit may influence the contribution of others. In this case, DD must explore complements at increasing levels of granularity through its recursive partition-and-reduction, as shown in Algorithm~\ref{alg:dd}.

If units are non-interacting, each unit contributes independently to the final output. In this case, the decision to discard a unit can be made by a single direct test: mask the unit and check whether the prediction changes. Consequently, the DD algorithm can be optimized into a one-pass procedure where each unit is evaluated only once, i.e., no combinations of units need to be evaluated.

For interacting units, the standard DD algorithm has the usual bounds:
$O(M \log M)$ in favorable cases and up to $O(M^2)$ in the worst case when fine
partitions are required. When the minimal sufficient set is small, DD often
eliminates large groups of units early. For non-interacting units, direct
one-at-a-time testing reduces the worst-case complexity to $O(M)$, since each unit is evaluated exactly once.

\begin{algorithm}[t]
\caption{Delta Debugging for Minimal Sufficient Sets}
\label{alg:dd}
\begin{algorithmic}[1]
\Statex \textbf{Input:} Units $U=\{u_1,\dots,u_M\}$, remainder network $f_{\mathrm{rem}}$
\Statex \textbf{Output:} 1-minimal sufficient set $S^\star$
\State $\hat{c} \gets \arg\max_c f_{\mathrm{rem}}(U)$
\State \Return \Call{DD}{$U, 2$}
\Function{DD}{$S, n$}
    \State Partition $S$ into $n$ subsets $S_1,\dots,S_n$ (sizes differ by at most one)
    \State $removed \gets \text{False}$
    \For{each $S_i$}
        \State $S_{\text{test}} \gets S \setminus S_i$
        \State Create $U'$ by zeroing all units not in $S_{\text{test}}$
        \If{$\arg\max_c f_{\mathrm{rem}}(U') = \hat{c}$}
            \State $removed \gets \text{True}$
            \State \Return \Call{DD}{$S_{\text{test}}, 2$}
        \EndIf
    \EndFor
    \State $n_{\text{next}} \gets \min(2n, |S|)$
    \If{\textbf{not} $removed$ \textbf{and} $n_{\text{next}} = n$}
        \State \Return S
    \EndIf
    \State \Return \Call{DD}{$S, n_{\text{next}}$}
\EndFunction
\end{algorithmic}
\end{algorithm}

\textbf{Illustrative Example.}
Suppose $S=\{1,\dots,8\}$, where units are interacting, and only units $2$ and $4$ are required to preserve the prediction. DD begins with $n=2$ and partitions $S$ into $\{1,2,3,4\}$ and $\{5,6,7,8\}$. Masking the first subset removes required units and changes the prediction, but masking the second preserves it, so $\{5,6,7,8\}$ is discarded and DD restarts with $S=\{1,2,3,4\}$ and $n=2$. Masking either half ($\{1,2\}$ or $\{3,4\}$) now removes a required unit, so DD increases the granularity to $n=4$ and tests singletons. Masking unit $1$ preserves the prediction, so it is removed and DD restarts with $S=\{2,3,4\}$. DD then partitions $\{2,3,4\}$ into $\{2,3\}$ and $\{4\}$, but masking either subset changes the prediction. Increasing to $n=3$ allows singleton testing again; masking unit $3$ preserves the prediction, so it is removed and DD restarts with $S=\{2,4\}$. With $S=\{2,4\}$, masking either unit changes the prediction. Since no subset can be removed and the granularity cannot increase further (i.e., $n = |S|$), the algorithm reaches a fixed point and returns the 1-minimal sufficient set $S^\star=\{2,4\}$.

\subsection{Step 3: Saliency Map Generation}
\label{subsec:saliencymap_generation}
Once a minimal subset of unit indices $S^\star$ is identified by delta debugging, we compute importance weights for each unit in $S^\star$ based on the drop in the logit for the predicted class $\hat{c}$ when that unit is removed. Let $y_{\hat{c}}$ be the original logit for class $\hat{c}$. For each unit index $i \in S^\star$, we 
mask the corresponding unit (feature map $A_i$ for CNNs or patch token $P_i$ for ViTs) while keeping the other units in $S^\star$ active, perform a partial 
forward pass to obtain the modified logit $y'_{\hat{c},i}$, and compute the drop 
$\delta_i = y_{\hat{c}} - y'_{\hat{c},i}$. The weight $w_i$ is then set to $\delta_i$ 
normalized such that $\sum_{i \in S^\star} w_i = 1$:
\[
w_i = \frac{\delta_i}{\sum_{j \in S^\star} \delta_j}
\]

This weighting strategy is similar to Ablation CAM~\cite{desai2020ablation}, as both derive weights from the effect of zeroing out individual units, but differs 
in normalization. We normalize across the selected minimal subset rather than by 
the original logit. This ensures that weights reflect the relative importance of 
units within the identified minimal set.

An interpretable saliency map $F_{exp}$ is generated by computing the weighted average of the selected units and upsampling to the original input image resolution. 
For CNNs:
\vspace{-0.6em}
\[
F_{exp} = \text{Upsample} \left( \text{Normalize} \left( \sum_{k \in S^\star} w_k A_k \right) \right)
\]

For ViTs, the weighted patch tokens are reshaped to the spatial grid and upsampled:
\vspace{-0.6em}

\[
F_{exp} = \text{Upsample} \left( \text{Normalize} \left( \text{Reshape} \left( \sum_{n \in S^\star} w_n P_n \right) \right) \right)
\]

Here, normalization is to min-max scaling, which maps the values to $[0,1]$. 
Bilinear interpolation is commonly used for upsampling. The resulting saliency map 
$F_{exp}$ highlights the spatial regions most critical for the model's prediction 
$\hat{c}$, with intensities proportional to unit importance.

\begin{table*}[!htbp]
\centering
\small
\caption{RQ1 Results: Average performance across CNNs with linear classifier heads (ResNet-50, EfficientNet-B0, Inception-v3), CNNs with non-linear classifier heads (VGG-11, VGG-16, VGG-19), and Vision Transformers (ViT-B/16, ViT-L/16). For each metric, we highlight the best method in bold: higher values are better for ADCC, Coh, Inc, and ADD; lower values are better for AD, Com, and Time.}
\begin{tabular}{llccccccc}
\toprule
\textbf{Group} & \textbf{Approach} & \textbf{ADCC↑} & \textbf{AD↓} & \textbf{Coh↑} & \textbf{Com↓} & \textbf{Inc↑} & \textbf{ADD↑} & \textbf{Time↓} \\
\midrule

\multirow{7}{*}{CNN Linear Heads}
& DD-CAM      & \textbf{0.8087} & \textbf{0.1393} & \textbf{0.9877} & \textbf{0.2137} & \textbf{0.4673} & \textbf{0.3873} & 1.5443 \\

& Grad-CAM      & 0.8010 & 0.1530 & 0.9790 & 0.2517 & 0.4200 & 0.3823 & 0.0240 \\

& Grad-CAM++    & 0.7727 & 0.1973 & 0.9857 & 0.2517 & 0.3590 & 0.3747 & 0.0290 \\

& XGrad-CAM     & 0.8010 & 0.1537 & 0.9790 & 0.2517 & 0.3740 & 0.3823 & \textbf{ 0.0140} \\

& Layer-CAM     & 0.7653 & 0.1980 & 0.9807 & 0.2450 & 0.3463 & 0.3500 & 0.0540 \\

& Score-CAM     & 0.7937 & 0.1653 & 0.9450 & 0.2300 & 0.4370 & 0.3560 & 1.6460 \\

& Ablation-CAM  & 0.7803 & 0.1920 & 0.9580 & 0.2197 & 0.4023 & 0.3627 & 1.5463 \\

& Recipro-CAM  & 0.7649 & 0.1510 & 0.9680 & 0.2246 & 0.4254 & 0.3715 & 0.0145 \\
\midrule

\multirow{7}{*}{CNN Non-Linear Heads}
& DD-CAM         & \textbf{0.7873} & \textbf{0.1610} & \textbf{0.9827} & 0.2170 & \textbf{0.3120} & \textbf{0.5423} & 1.8267 \\
& Grad-CAM       & 0.7437 & 0.1913 & 0.9437 & 0.2567 & 0.2790 & 0.5120 & 0.0110 \\

& Grad-CAM++     & 0.7303 & 0.2077 & 0.9737 & 0.2373 & 0.2727 & 0.5217 & 0.0210 \\

& XGrad-CAM      & 0.7500 & 0.1927 & 0.9500 & 0.2547 & 0.3113 & 0.5110 & \textbf{0.0107} \\

& Layer-CAM      & 0.7033 & 0.2993 & 0.9643 & \textbf{0.1880} & 0.2843 & 0.5070 & 0.1347 \\

& Score-CAM      & 0.7740 & 0.1677 & 0.9450 & 0.2383 & 0.2560 & 0.5223 & 1.2117 \\

& Ablation-CAM   & 0.7507 & 0.2413 & 0.9707 & 0.2253 & 0.2820 & 0.5397 & 1.1983 \\

& Recipro-CAM  & 0.7754 & 0.1735 & 0.9755 & 0.2815 & 0.2929 & 0.4813 & 0.0184 \\
\midrule

\multirow{8}{*}{Vision Transformers}
& DD-CAM     & 0.7985 & \textbf{0.0832}  & 0.\textbf{9950} & 0.2010  & \textbf{0.6000} & \textbf{0.3975} & 1.9301 \\

& Grad-CAM      & 0.1721 & 0.7851 & 0.7244 & 0.2303 & 0.0929 & 0.1147 & 0.0415 \\

& Grad-CAM++    & 0.1345 & 0.8563 & 0.7479 & 0.1498 & 0.1000 & 0.0793 & 0.0260 \\

& XGrad-CAM     & 0.2190 & 0.8611 & 0.7331 & \textbf{0.1182} & 0.0200 & 0.0549 & \textbf{0.0238} \\

& Layer-CAM     & 0.1831 & 0.9468 & 0.7704 & 0.1930 & 0.0150 & 0.1491 & 0.0273 \\

& ScoreCAM         & \textbf{0.8103} & 0.1529 & 0.8641 & 0.2883 & 0.2250 & 0.2028 & 4.0864 \\

& Ablation-CAM  & 0.4389 & 0.5615 & 0.8280 & 0.1618 & 0.3000 & 0.1062 & 6.9294 \\

& Recipro-CAM   & 0.6889 & 0.1381 & 0.9850 & 0.5351 & 0.4250 & 0.3954 & 0.0704 \\
\bottomrule
\end{tabular}
\end{table*}

\begin{table}
\centering
\small
\caption{RQ2 Results: Average localization performance on 1,000 chest X-rays (DenseNet-121). Best values are in bold (higher is better for IoU, Precision, Recall; lower for Regions).}
\label{tab:medical_localization}
\begin{tabular}{l c c c c}
\toprule
\textbf{Approach} & \textbf{IoU $\uparrow$} & \textbf{Precision $\uparrow$} & \textbf{Recall $\uparrow$} & \textbf{Regions $\downarrow$} \\
\midrule
\textbf{DD-CAM} & \textbf{0.263} & \textbf{0.307} & \textbf{0.692} & \textbf{1.00} \\
GradCAM                & 0.060 & 0.088 & 0.204 & 1.41 \\
GradCAM++              & 0.135 & 0.171 & 0.519 & 1.11 \\
XGrad-CAM              & 0.049 & 0.081 & 0.161 & 1.39 \\
LayerCAM               & 0.137 & 0.183 & 0.567 & 1.10 \\
Score-CAM              & 0.055 & 0.093 & 0.149 & 1.02 \\
Ablation-CAM           & 0.059 & 0.086 & 0.191 & 1.32 \\
Recipro-CAM            & 0.181 & 0.251 & 0.568 & 1.37 \\
\bottomrule
\end{tabular}
\end{table}

\begin{figure*}[t]
\centering
\begin{subfigure}{\textwidth}
    \centering
    \includegraphics[width=\textwidth]{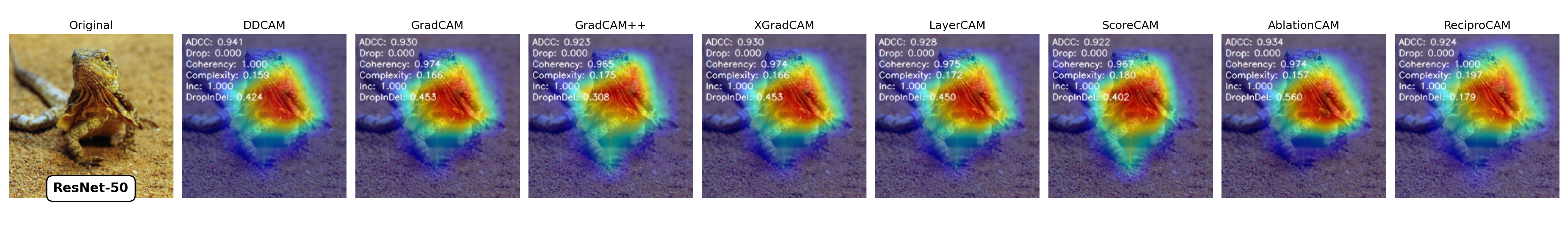}

\end{subfigure}\vspace{-0.6em}
\begin{subfigure}{\textwidth}
    \centering
    \includegraphics[width=\textwidth]{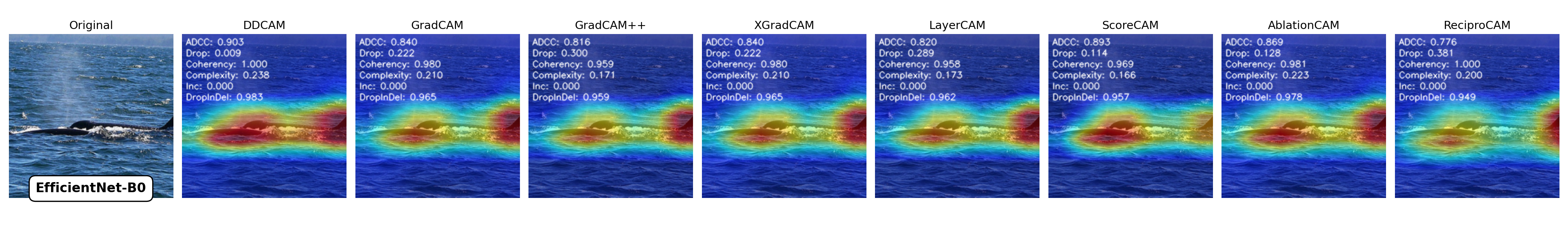}

\end{subfigure}\vspace{-0.6em}
\begin{subfigure}{\textwidth}
    \centering
    \includegraphics[width=\textwidth]{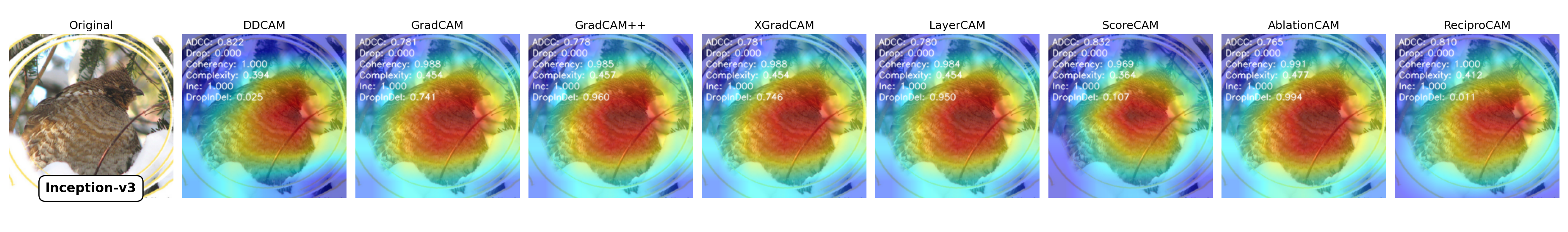}
\end{subfigure}\vspace{-0.6em}

\begin{subfigure}{\textwidth}
    \centering
    \includegraphics[width=\textwidth]{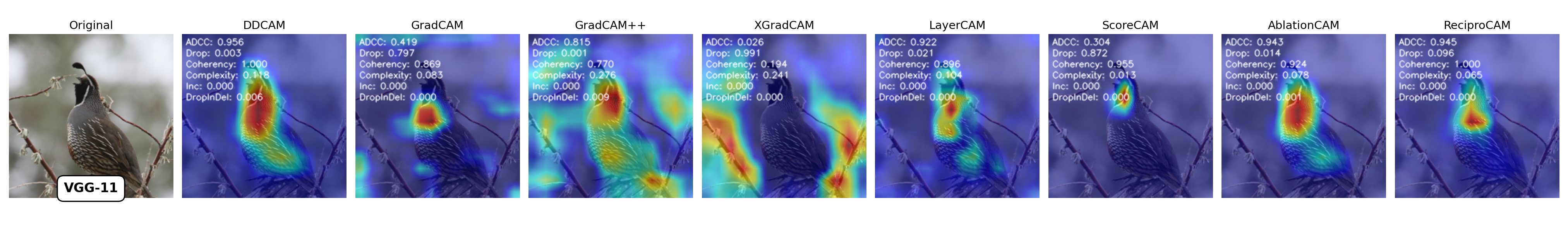}

\end{subfigure}\vspace{-0.6em}
\begin{subfigure}{\textwidth}
    \centering
    \includegraphics[width=\textwidth]{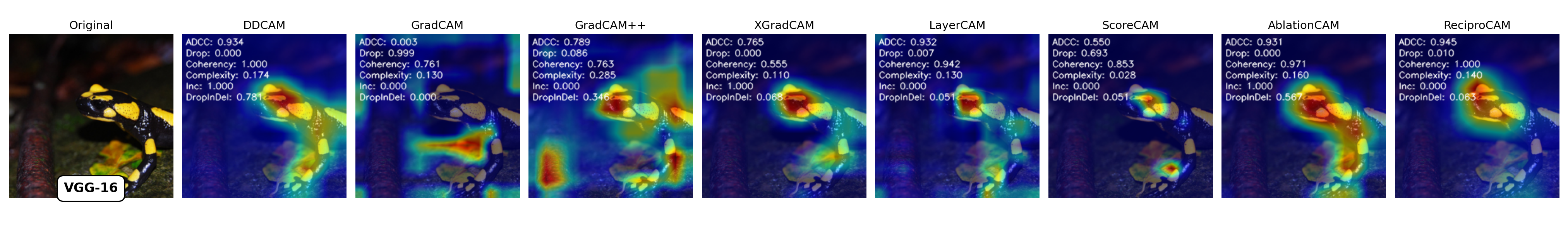}

\end{subfigure}\vspace{-0.6em}

\begin{subfigure}{\textwidth}
    \centering
    \includegraphics[width=\textwidth]{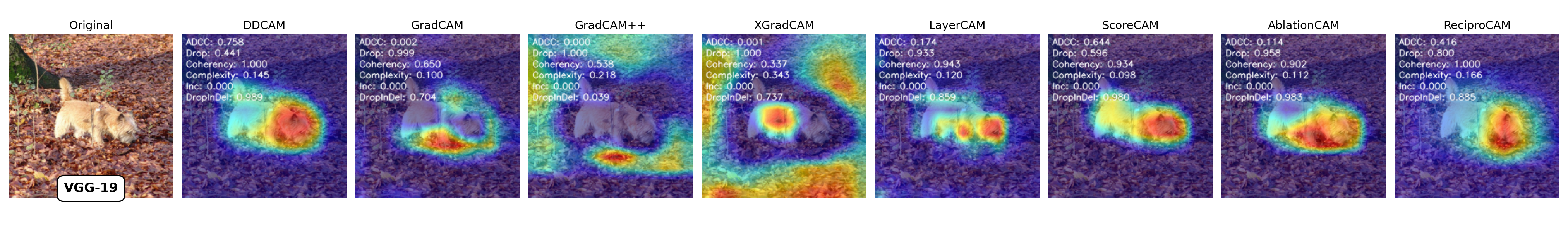}

\end{subfigure}\vspace{-0.6em}

\begin{subfigure}{\textwidth}
    \centering
    \includegraphics[width=\textwidth]{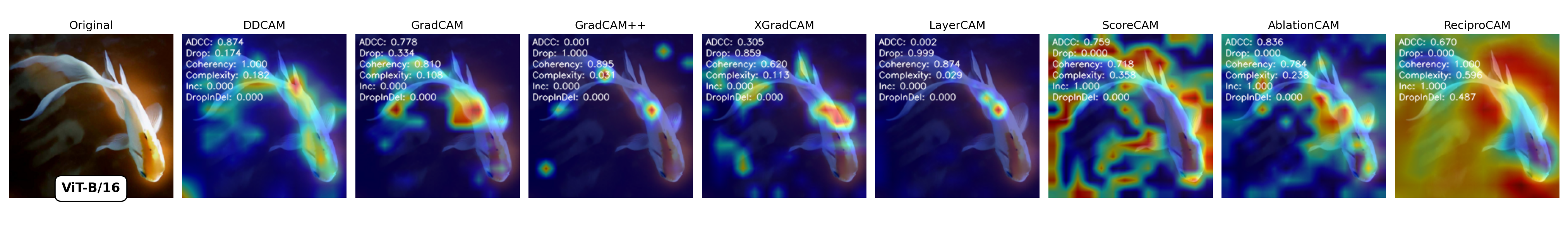}
  
\end{subfigure}\vspace{-0.6em}

\begin{subfigure}{\textwidth}
    \centering
    \includegraphics[width=\textwidth]{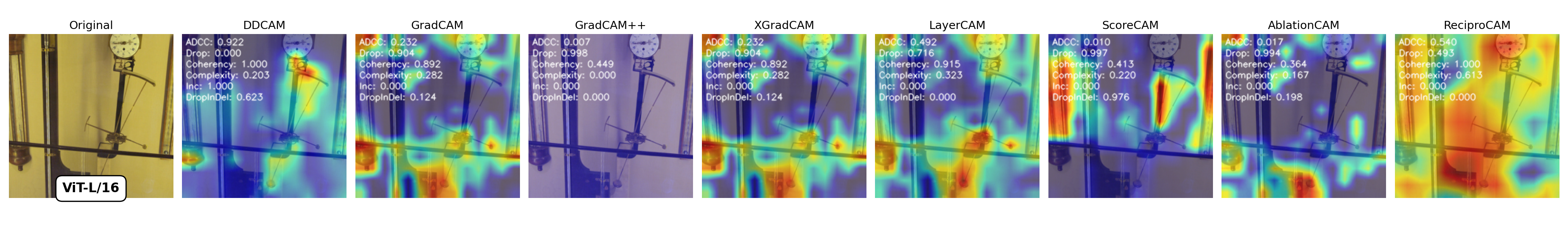}
  
\end{subfigure}\vspace{-0.8em}
\caption{Qualitative comparison of saliency maps generated by our approach (DD-CAM) and baseline approaches (Grad-CAM, Grad-CAM++, XGrad-CAM, Layer-CAM, Score-CAM, Ablation-CAM, and Recipro-CAM).}
\label{fig:qualitative_examples}
\end{figure*}

\begin{figure*}[t]
\centering

\begin{subfigure}[b]{1\textwidth}
    \centering
    \includegraphics[width=\textwidth]{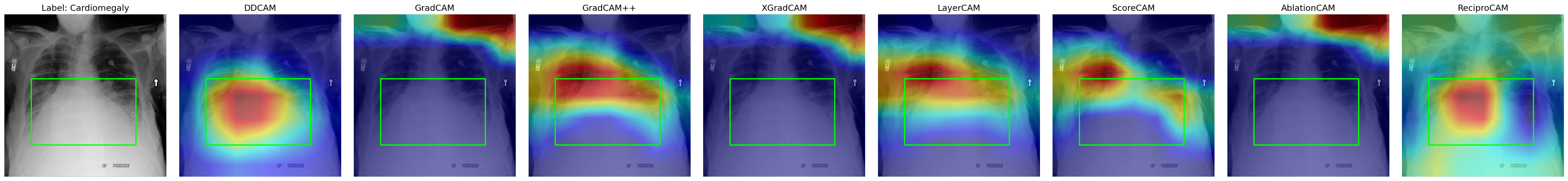}
\end{subfigure}\vspace{-0.5em}

\begin{subfigure}[b]{1\textwidth}
    \centering
    \includegraphics[width=\textwidth]{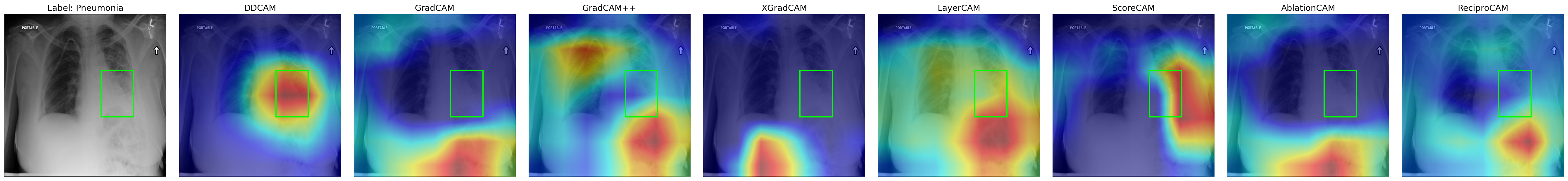}
\end{subfigure}\vspace{-0.8em}

\caption{Qualitative localization examples on NIH ChestX-ray14. DD-CAM consistently isolates a single, compact pathological region while baselines vary between diffuse and fragmented responses.}
\label{fig:medical_examples}
\end{figure*}


\section{Experimental Design}
\label{sec:exp-design}
We evaluate our approach on vision models to answer two key research questions:

\textbf{RQ1 (Faithfulness):} How do the generated saliency maps reflect the model’s decision-making process, and how does our approach compare with state-of-the-art CAM methods across architectures?

\textbf{RQ2 (Localization):} How do saliency maps align with human-annotated regions of interest, and how does our approach compare with state-of-the-art methods in localization accuracy?

\subsection{Datasets}
For RQ1, we use 2{,}000 randomly sampled images from the ILSVRC2012 validation set~\cite{russakovsky2015imagenet}, resized to $224\times224$ and normalized following standard ImageNet preprocessing. For RQ2, we use 1{,}000 chest radiographs from NIH ChestX-ray14~\cite{wang2017chestx} together with the radiologist-annotated bounding boxes provided by the dataset. This allows the assessment of how well the saliency maps align with clinically relevant regions identified by human experts.

\subsection{Models}
We evaluate eight architectures spanning convolutional and transformer paradigms.

\textbf{CNNs with linear classifier heads.}
ResNet-50~\cite{he2016deep}, EfficientNet-B0~\cite{tan2019efficientnet}, and Inception-v3~\cite{szegedy2016rethinking} use global average pooling followed by a single fully connected layer, where each feature map contributes additively to the final prediction. For such models, the representational units are linearly independent, allowing an optimized version of delta debugging. When contributions are additive, each unit can be tested once to determine whether it is necessary for preserving the prediction, reducing the worst-case complexity from $O(n^2)$ to $O(n)$. This optimization is implemented by (a) fixing the granularity to $n_{\text{init}} = K$ (each unit tested individually) and (b) proceeding sequentially without restarting after each removal.

\textbf{CNNs with non-linear classifier heads.}
VGG-11, VGG-16, and VGG-19~\cite{simonyan2014very} contain multiple fully connected layers with ReLU activations, introducing nonlinear dependencies among the final-layer feature maps. For these interacting units, we employ the standard delta debugging procedure, starting with coarse subdivisions ($n_{\text{init}} = 2$) and recursively refining the granularity as needed.

\textbf{Vision Transformers.}
ViT-B/16 and ViT-L/16~\cite{dosovitskiy2020image} divide images into $16 \times 16$ patches, producing 196 patch tokens plus one CLS token. Although the classifier head is linear, the CLS token is computed after the final transformer block, where all patches interact through multi-head self-attention. Consequently, ViTs exhibit interacting units for sufficiency testing, and we configure delta debugging with coarse initial granularity ($n_{\text{init}} = 2$), refining it recursively as needed.

All models use TorchVision pretrained weights~\cite{torchvision2016}.

\textbf{Medical Imaging Model.}  
For the ChestX-ray14 localization experiments (RQ2), we use a DenseNet-121 model pretrained on large-scale chest X-ray datasets~\cite{cohen2022torchxrayvision}.

\subsection{Baselines}
We compare our approach to six commonly used CAM-based attribution approaches: Grad-CAM~\cite{selvaraju2017grad}, Grad-CAM++~\cite{chattopadhay2018grad}, XGrad-CAM~\cite{fu2020axiom}, Layer-CAM~\cite{jiang2021layercam}, Ablation-CAM~\cite{desai2020ablation}, and Score-CAM~\cite{wang2020score}. These approaches differ in how they estimate feature importance—via gradients, multi-layer aggregation, ablations, or activation perturbations—but all operate on the final-layer representation and therefore provide a consistent comparison framework. For ViTs, all baselines operate on patch tokens, following standard practice. Implementations are based on the PyTorch-CAM library~\cite{jacobgilpytorchcam}. We restrict comparisons to CAM-style approaches because perturbation-based metrics are not directly compatible with input-level approaches such as LIME~\cite{ribeiro2016should} or SHAP~\cite{lundberg2017unified}.

\subsection{Evaluation Metrics}
For RQ1, we evaluate faithfulness on ImageNet using six standard perturbation-based metrics~\cite{chattopadhay2018grad, jung2021towards, poppi2021revisiting}: 
\emph{Average Drop (AD, $\downarrow$)} quantifies the confidence decrease when replacing 
the input with the saliency map; \emph{Coherency (Coh, $\uparrow$)} measures consistency 
between full-image and saliency map confidences; \emph{Complexity (Com, $\downarrow$)} 
captures explanation sparsity via the L1 norm of the normalized heatmap; 
\emph{Average DCC (ADCC, $\uparrow$)} summarizes AD, Coh, and Com as a harmonic mean; 
\emph{Increase in Confidence (IC, $\uparrow$)} counts cases where the saliency map 
increases confidence; and \emph{Average Drop in Deletion (ADD, $\uparrow$)} measures confidence decreases when masking highlighted regions, reflecting feature criticality.

For RQ2, we evaluate localization quality against radiologist-annotated bounding boxes using \emph{IoU ($\uparrow$)}, \emph{Precision ($\uparrow$)}, 
\emph{Recall ($\uparrow$)}, \emph{Percentage Highlighted ($\downarrow$)}, and \emph{Number of Regions ($\downarrow$)} ~\cite{everingham2010pascal, desai2020ablation, wang2020score}. IoU, precision, and recall measure region overlap, correctness of highlighted pixels, and coverage of the pathology, respectively, while the latter two quantify explanation focus and spatial coherence. All experiments use PyTorch on a 12 GB GPU system with 32 GB RAM, and results are averaged over five runs.

\section{Results and Discussion}
\label{sec:results}
We evaluate DD-CAM against state-of-the-art CAM-based approaches to assess faithfulness and localization quality.

\subsection{Faithfulness Evaluation (RQ1)}
Across all model families, DD-CAM produces the most faithful and compact saliency maps. Averaged over three CNNs with linear heads, DD-CAM achieves the highest ADCC (0.8087), lowest Average Drop (0.1393), highest Coherency (0.9877), and lowest Complexity (0.2137), indicating that the highlighted regions both preserve model confidence and remain highly focused. Similar trends appear
for nonlinear CNNs, where DD-CAM again achieves the highest ADCC (0.7873), lowest AD (0.1610), and strongest causal relevance (ADD = 0.5423), despite the feature interactions introduced by nonlinear classifier heads. Layer-CAM achieves a slightly lower Complexity (0.1880) in this group, though at the cost of reduced causal faithfulness (lower ADCC and higher Average Drop).

For Vision Transformers, DD-CAM achieves the lowest AD (0.0832), highest Increase in Confidence (0.600), and highest ADD (0.3975), showing that the regions corresponding to the minimal token set remain essential despite self-attention mixing. Score-CAM yields a slightly higher ADCC (0.8103) but produces substantially larger regions and requires 2.4× more computation, resulting in worse sparsity and efficiency. XGrad-CAM attains the lowest Complexity (0.1182) but exhibits weak causal alignment, as shown by its lower ADCC and higher Average Drop values.

In terms of runtime, gradient-based (e.g., Grad-CAM, XGrad-CAM) approaches are faster. DD-CAM remains efficient among gradient-free approaches like Score-CAM and Ablation-CAM by operating only on the final representational layer instead of requiring full forward pass.

Figure~\ref{fig:qualitative_examples} shows saliency map comparisons for representative samples, illustrating how DD-CAM’s minimal decision-preserving unit selection reduces clutter and yields clearer, more focused visualizations.

\subsection{Localization (RQ2)}
\label{sec:medical_results}
On ChestX-ray14, DD-CAM produces regions that closely align with
radiologist-annotated regions. As shown in Table~\ref{tab:medical_localization}, it achieves the highest IoU (0.263), precision (0.307), and recall (0.692), outperforming the strongest baseline by approximately +45\%, +22\%, and +22\%, respectively. Relative to the widely used Grad-CAM, these gains are even more pronounced (4.4$\times$ in IoU and 3.5$\times$ in precision), indicating substantially improved localization fidelity.

DD-CAM also yields the most compact explanations, averaging only 1.00 region per image, whereas baselines highlight 1.02–1.41 regions. This reduced fragmentation is evident in Figure~\ref{fig:medical_examples}, where DD-CAM isolates a single coherent pathological focus, while other methods yield diffuse or fragmented activations. Overall, DD-CAM provides the most human-aligned and semantically meaningful localizations among CAM-based methods.








\section{Limitations}
Although DD-CAM mitigates key limitations of CAM-based explanations—most notably by selecting minimal, decision-preserving feature map subsets—it still shares some constraints of the CAM paradigm. Like other CAM-based approaches, DD-CAM saliency maps by upsampling the selected feature maps to match the input resolution. This post-processing step can introduce spatial imprecision, potentially affecting the granularity of highlighted regions. We mitigate this by focusing only on the most critical feature maps. Furthermore, our approach is inherently white-box, requiring access to internal activations and weights.

\section{Conclusion}
\label{sec:conclusion}
We introduce the first approach that applies the principle of minimal sufficiency to internal representations to produce minimal sufficient explanations for vision models in a representation-agnostic manner. Specifically, we ask which subset of representational units (feature maps for CNNs, patch tokens for ViTs) is jointly sufficient to preserve the model’s predictions. By adapting software debugging strategies, the framework provides formal 1-minimality guarantees for the resulting explanations, ensuring that no redundant representational units remain. Empirically, DD-CAM improves faithfulness across the majority of model–metric combinations and enhances localization. \textbf{Future directions:} (1) Using minimal unit sets for model debugging and bias analysis; (2) Extending the framework to other domains by redefining units.

\section*{Acknowledgment}
This work is supported by a research grant (70NANB21H092) from the Information Technology Laboratory of the National Institute of Standards and Technology (NIST).

Disclaimer: Certain equipment, instruments, software, or materials are identified in this paper in order to specify the experimental procedure adequately.  Such identification is not intended to imply recommendation or endorsement of any product or service by NIST, nor is it intended to imply that the materials or equipment identified are necessarily the best available for the purpose.

\small
\bibliographystyle{ieeenat_fullname}
\bibliography{main}


\end{document}